# Fuel Efficiency Analysis of the Public Transportation System Based on the Gaussian Mixture Model Clustering

Zhipeng Ma[1][0000-0002-4049-539X], Bo Nørregaard Jørgensen[1][0000-0001-5678-6602] and Zheng Ma[1][0000-0002-9134-1032]

[1]SDU Center for Energy Informatics, Maersk Mc-Kinney Moller Institute, Faculty of Engineering, University of Southern Denmark, DK-5230 Odense, Denmark
{zhma, bnj, zma}@mmmi.sdu.dk

**Abstract.** Public transportation is a major source of greenhouse gas emissions, highlighting the need to improve bus fuel efficiency. Clustering algorithms assist in analyzing fuel efficiency by grouping data into clusters, but irrelevant features may complicate the analysis and choosing the optimal number of clusters remains a challenging task. Therefore, this paper employs the Gaussian mixture models to cluster the solo fuel-efficiency dataset. Moreover, an integration method that combines the Silhouette index, Calinski-Harabasz index, and Davies-Bouldin index is developed to select the optimal cluster numbers. A dataset with 4006 bus trips in North Jutland, Denmark is utilized as the case study. Trips are first split into three groups, then one group is divided further, resulting in four categories: extreme, normal, low, and extremely low fuel efficiency. A preliminary study using visualization analysis is conducted to investigate how driving behaviors and route conditions affect fuel efficiency. The results indicate that both individual driving habits and route characteristics have a significant influence on fuel efficiency.

**Keywords:** Fuel efficiency, Clustering, Gaussian mixture models, Statistical analysis.

## 1 Introduction

Anthropogenic greenhouse gases (GHG) have given rise to global warming through the greenhouse effect in recent years, resulting in numerous ecological and social problems. The transportation sector contributes to almost 25% of GHG emissions in the European Union (EU) [1], about 76% of which are caused by road transport [2]. To achieve the United Nations' sustainability goals, eco-driving strategies need to be investigated aimed at reducing fuel consumption and CO2 emissions.

Public transportation, particularly bus networks, plays a significant role in urban mobility, providing a sustainable alternative to individual vehicle usage. Therefore, bus networks represent a significant contributor to energy usage and GHG emissions. Enhancing the fuel efficiency of the systems holds promise for reducing negative environmental impact and enhancing sustainability in urban environments [3-4].

To enhance the fuel efficiency of buses, it is essential to analyze the influencing factors regarding driving behaviors and route conditions [5]. Before commencing the feature relationship analysis, it is imperative to categorize the bus trips into distinct groups, as their impact on ecological driving and low fuel efficiency driving varies significantly. The clustering algorithm serves as a powerful tool to categorize bus trips based on distinct fuel consumption patterns. Spectral clustering is employed in [6] to group drivers and measure the effect of their driving behavior on fuel consumption. The K-means and K-medoids algorithms are utilized in [7] to identify representative urban driving cycles for estimating fuel consumption and emission rates. Additionally, a novel clustering algorithm based on a weighted correlation mechanism is developed in [8] to categorize vehicles according to the features related to $CO_2$ emissions.

In general, the existing frameworks employ clustering algorithms to categorize multivariate datasets. However, this approach might introduce noise into the fuel efficiency analysis, as not all driving and route patterns have a direct impact on fuel efficiency. Moreover, selecting the cluster's number is still challenging when applying the clustering algorithms. To address the mentioned research gap, this paper utilizes the Gaussian mixture model (GMM) clustering algorithm [9] to categorize solo fuel efficiency data, and employs an integration method to select the cluster's number. The trips with different fuel efficiency are categorized into various clusters, aiming at analyzing the influencing factors of different levels of fuel efficiency in future research.

A dataset with 4006 bus trips in North Jutland, Denmark is utilized as the case study. Firstly, an integration method combines the Silhouette index (SI) [10], the Calinski-Harabasz index (CHI) [11], and the Davies-Bouldin index (DBI) [12] to evaluate the performance of the clustering with varying numbers of clusters, aiming at selecting the optimal number of clusters. The trips are initially categorized into three groups, with one of the groups further divided into two based on the data distributions. This results in a total of four groups, including extreme fuel efficiency, normal fuel efficiency, low fuel efficiency, and extremely low fuel efficiency. Afterward, a preliminary investigation is conducted to explain the impact of driving behaviors and route conditions on fuel efficiency based on bar chart analysis. The findings reveal that individual driving habits and route characteristics have a significant influence on fuel efficiency.

The remainder of the paper is structured as follows. Section 2 introduces the methodology of the proposed framework. Section 3 clarifies the dataset in the case study and section 4 highlights the results and findings. Section 5 discusses the results and introduces future work. Lastly, Section 6 concludes the findings of the study.

## 2      Methodology

The paper presents a three-step analysis of fuel efficiency in public transport. First, it begins with the collection of fuel-efficiency data. Second, it employs multiple clustering evaluation indexes to determine the optimal number of clusters in Gaussian Mixture Model (GMM) clustering, and then refines these clusters through statistical analysis. Finally, it conducts a detailed statistical analysis of each cluster's distribution and examines the factors affecting fuel efficiency.

### 2.1 GMM Clustering Algorithm

The GMM clustering algorithm is chosen for its flexibility in accommodating mixed membership and varying cluster shapes, unlike the spherical clusters of K-means [13]. It uses a probabilistic model assuming data points are from multiple Gaussian distributions, aiming to uncover these distributions and assign points to them. The probability density function of the multivariate GMM is:

$$p(x) = \sum_{i=1}^{K} \pi_i \mathcal{N}(x|\boldsymbol{\mu}_i, \boldsymbol{\Sigma}_i) \tag{1}$$

where $K$ refers to the number of mixture components, $\pi_i$ represents the weight of the $i$-th Gaussian component, and $\mathcal{N}(x|\boldsymbol{\mu}_i, \boldsymbol{\Sigma}_i)$ denotes the $i$-th Gaussian distribution with the mean $\boldsymbol{\mu}_i$ and the covariance $\boldsymbol{\Sigma}_i$.

Parameters of the Gaussian process, like mean and covariance, are estimated using the EM algorithm, which iteratively adjusts parameters through an E-step, calculating the expected log-likelihood, and an M-step, maximizing this expectation. For a GMM with K components and N observations, the E-step softly assigns each sample to a component.

$$\gamma_j(\mathrm{x}_n) \leftarrow \frac{\pi_j \mathcal{N}(\mathrm{x}_n|\boldsymbol{\mu}_j, \boldsymbol{\Sigma}_j)}{\sum_{k=1}^{K} \pi_k \mathcal{N}(\mathrm{x}_n|\boldsymbol{\mu}_k, \boldsymbol{\Sigma}_k)} \quad \forall j = 1, \ldots K; \; n = 1, \ldots, N \tag{2}$$

where $\gamma_j(\mathrm{x}_n)$ refers to the probability that the $n$-th data point is generated by the $j$-th Gaussian component and $\pi_j$ denotes the weight of the $j$-th component, where $0 \leq \pi_j \leq 1$ for each $j \in \{1, \ldots K\}$ and $\sum_{k=1}^{K} = 1$.

In the M-step, the parameters $\pi_j$, $\boldsymbol{\mu}_j$ and $\boldsymbol{\Sigma}_j$ are re-estimated and updated separately for each mixture component based on the soft assignments in the E-step following the order of equations (3) to (6).

$$\widehat{N}_j \leftarrow \sum_{n=1}^{N} \gamma_j(\mathrm{x}_n) \tag{3}$$

$$\widehat{\pi}_j^{\text{new}} \leftarrow \frac{\widehat{N}_j}{N} \tag{4}$$

$$\widehat{\boldsymbol{\mu}}_j^{\text{new}} \leftarrow \frac{1}{\widehat{N}_j} \sum_{n=1}^{N} \gamma_j(\mathrm{x}_n) \mathrm{x}_n \tag{5}$$

$$\widehat{\boldsymbol{\Sigma}}_j^{\text{new}} \leftarrow \frac{1}{\widehat{N}_j} \sum_{n=1}^{N} \gamma_j(\mathrm{x}_n) (\mathrm{x}_n - \widehat{\boldsymbol{\mu}}_j^{\text{new}})(\mathrm{x}_n - \widehat{\boldsymbol{\mu}}_j^{\text{new}})^{\mathrm{T}} \tag{6}$$

### 2.2 Selection of Cluster Numbers

The initial determination of the number of clusters is crucial when applying the GMM clustering algorithm. Both cohesion and separation metrics need to be considered to select the optimal number of clusters [14]. Cohesion denotes the degree of proximity among data points within a cluster, while separation signifies the distinctiveness between clusters. In this case, three clustering evaluation indexes, including the Silhouette index (SI) [10], the Calinski-Harabasz index (CHI) [11], and the Davies-Bouldin index (DBI) [12], are employed to assess the performance of the clustering with varying numbers of clusters. Subsequently, the ranking of each index is calculated and the average of these rankings is utilized as the coefficient to select the optimal number of clusters.

**The Silhouette index (SI).** The SI considers the pairwise distance within the between clusters. It ranges between -1 and 1, and an SI close to 1 indicates a good partition of data points.

For the $i$-th sample in the $I$-th cluster $C_I$, $I = 1, ..., K$, the silhouette width is:

$$s_i = \frac{b_i - a_i}{\max(a_i, b_i)} \tag{7}$$

where $a_i$ refers to the mean distance between the $i$-th instance and all other samples in $C_I$, and $b_i$ denotes the smallest mean distance between the $i$-th instance in $C_I$ and all samples outside $C_I$.

Assuming the number of data points in the cluster $C_I$ is $M$, the SI for $C_I$ is:

$$SI_I = \frac{1}{M} \sum_{i=1}^{M} s_i \tag{8}$$

The SI for the entire dataset with $K$ clusters is:

$$SI = \max_{K} SI_I \tag{9}$$

**The Calinski-Harabasz Index (CHI).** The CHI calculates the sum of squares of values between and within the clusters. A higher value of the CHI represents a better performance of clustering. Assuming a dataset with $N$ observations is clustered into $K$ clusters, the CHI is:

$$CHI = \frac{BC}{WC} \frac{N-K}{K-1} \tag{10}$$

where $BC$ represents the weighted sum of squared Euclidean distances between each cluster centroid and the overall centroid of the dataset, and $WC$ refers to the sum of squared Euclidean distances between data points and their respective cluster centroids.

**The Davies-Bouldin index (DBI).** The DBI computes the average similarity between each cluster and its most similar cluster, where the similarity signifies the ratio of the within-cluster scatter to the between-cluster distance. A lower DBI value indicates a better performance of clustering. Assuming a dataset with N observations is clustered into K clusters, the DBI is:

$$DBI = \frac{1}{K} \sum_{i=1}^{K} \max_{j \neq i} \left( \frac{d_i + d_j}{d(c_i + c_j)} \right) \tag{11}$$

Where $c_i$ and $c_j$ are the centroids of clusters $i$ and $j$, $d_i$ and $d_j$ denote the average Euclidean distances from $c_i$ to the data points in the cluster $i$ and $c_j$ to the data points in the cluster $j$ respectively, and $d(c_i + c_j)$ refers to the Euclidean distance between $c_i$ and $c_j$.

## 3    Case Study

The proposed framework based on GMM clustering is employed to analyze the bus trips from the public transport system of North Jutland, Denmark. A bus trip in this study is defined as a traveled way from the first stop to the destination of a bus plan in a specific time. A total of 4006 trips are recorded in the dataset and the feature 'fuel efficiency' denotes the fuel consumption in liters per 100 kilometers (L/100km) according to the EU standard. The experiments are conducted in the Visual Studio Code platform (version: 1.85.2) and all codes are written in Python (version: 3.10.9).

Fig. 1 illustrates a bus route in the case study, where the green lines refer to the route and the green hollow circles denote bus stops. Fig. 2 visualizes the distribution of fuel efficiency using a histogram. The horizontal axis refers to the fuel efficiency of the samples and the vertical axis shows the number of samples in each group of the fuel efficiency. All 4006 samples are divided into 10 groups in Fig. 2, with each group of fuel efficiency represented by a light blue bar.

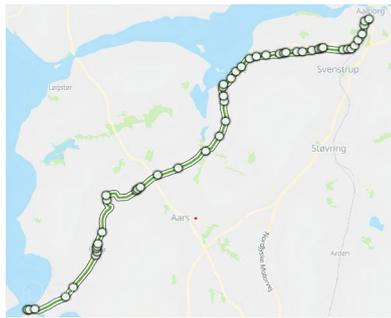 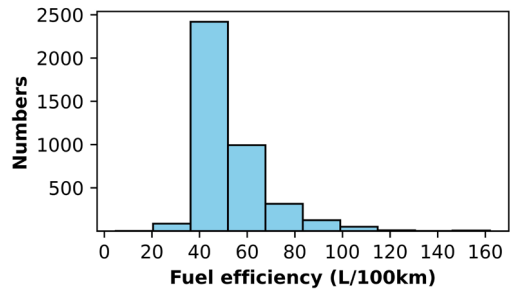

**Fig. 1.** Example of a bus route in the case study.

**Fig. 2.** The distribution of fuel efficiency of 4006 samples.

## 4    Results

### 4.1    GMM Clustering

First of all, the number of clusters is selected from the range [2, 9] based on the methodology outlined in Section 2.2. Table 1 demonstrates the ranking of clustering performance under different cluster numbers. The average ranking shows that utilizing 3 clusters yields the best performance, taking into account both cohesion and separation properties.

Table 1. The ranking of clustering performance under the different number of clusters.

| No. Clusters | 2 | 3 | 4 | 5 | 6 | 7 | 8 | 9 |
|---|---|---|---|---|---|---|---|---|
| **SI** | **1** | 2 | 7 | 5 | 8 | 3 | 4 | 6 |
| **CHI** | 2 | **1** | 3 | 4 | 6 | 5 | 8 | 7 |
| **DBI** | 2 | **1** | 4 | 3 | 8 | 6 | 5 | 7 |
| **Avg.** | 1.7 | **1.3** | 4.7 | 4.0 | 7.3 | 4.7 | 5.7 | 6.7 |

In the next step, a GMM model with three clusters is employed to cluster the 4006 fuel-efficiency samples. Fig. 3 illustrates the clustering results in both the Gaussian distribution curve of each cluster and the scatter points.

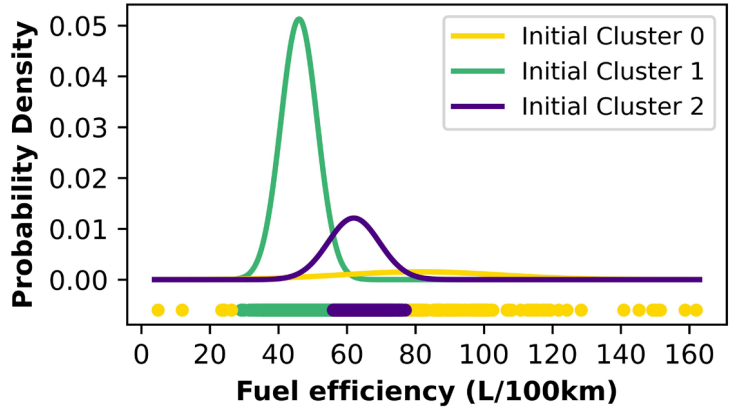

**Fig. 3.** Initial GMM clustering results.

As shown in Fig. 3, each Gaussian distribution curve represents the distribution of a cluster, and each data point (the scatter point under the curves) is assigned to the cluster yielding the highest posterior density. Therefore, the scatter points with the same color belong to the same cluster, represented by a Gaussian curve of that color. Cluster 0 (yellow one) represents the bias, including the extremely low and high fuel consumption. Cluster 1 (green one) and Cluster 2 (purple one) represents the normal and high fuel efficiency respectively. In the initial clustering, Cluster 0 exhibits a separable internal structure, suggesting a potential for further division into two subgroups. The data points in this cluster are completely partitioned into two parts by other clusters.

As displayed in Fig. 4, the optimized clustering analysis includes four clusters, where the initial Cluster 0 in Fig. 3 is divided into the new Cluster 0 and Cluster 3 in Fig. 4. The scatter points with the same colors are assigned to the same cluster shown in the legend. The new Cluster 0 (orange one) represents the extreme fuel efficiency, and the new Cluster 3 (brown one) represents the extremely low fuel efficiency.

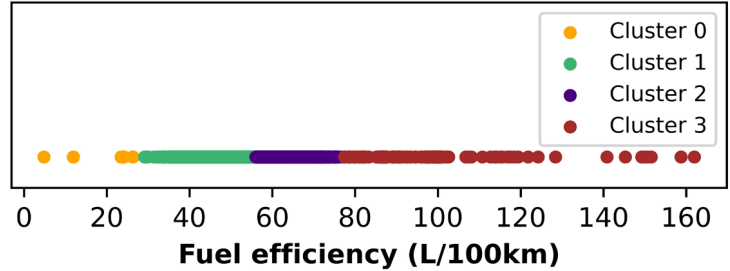

**Fig. 4.** Optimized GMM clustering results.

Table 2 shows the statistical values of each new cluster. The minimum and maximum values suggest that all data points are separated without overlapping in the ranges. Most data points are assigned to Cluster 1 and Cluster 2, indicating that the fuel efficiency of most trips is in the range of [29.16, 77.01] L/100km. However, the standard deviation of data points in Cluster 3 is significantly greater than those in other clusters, indicating the presence of outliers in Cluster 3.

**Table 2.** Statistical values of each cluster.

| Clusters | num | min | max | mean | median | std | definition |
|---|---|---|---|---|---|---|---|
| **Cluster 0** | 11 | 4.84 | 26.35 | 21.73 | 24.04 | 6.55 | Extremely fuel efficiency |
| **Cluster 1** | 2857 | 29.16 | 56.02 | 46.03 | 46.25 | 5.02 | Normal fuel efficiency |
| **Cluster 2** | 908 | 56.04 | 77.01 | 63.94 | 63.31 | 5.61 | Low fuel efficiency |
| **Cluster 3** | 230 | 77.44 | 161.94 | 95.05 | 90.86 | 14.79 | Extremely low fuel efficiency |

### 4.2 Influencing Factors of Fuel Consumption

This section conducts a brief analysis of whether individual driving behavior and route conditions significantly impact fuel consumption. The proportion of drivers' IDs and routes' IDs in each cluster is thoroughly analyzed in the following sub-sections through a detailed discussion of the bar charts shown in Fig. 5 and Fig. 6.

**Drivers.** This section examines the fuel-efficiency performance of 202 drivers in the recorded 4006 trips based on the 'DriverId' feature. The analysis categorizes the trips made by these drivers into four clusters in Section 4.1 to observe patterns in fuel efficiency. The proportion of each driver's trips across the four clusters is compared to the average proportion of all trips, to identify individual drivers' performance in terms of fuel efficiency.

Fig. 5 presents a bar chart depicting the proportion of drivers across four clusters. To better illustrate the results, the first 50 of the 202 drivers are selected as a subset for the bar chart. The x-axis lists the Drivers' IDs, representing individual drivers, and the y-axis indicates the percentage of each driver's trips in each cluster relative to their total trips. 'C0' – 'C3' in the legend denote 'Cluster 0' - 'Cluster 3' in Fig. 4 respectively. Each bar is divided into four segments, corresponding to the percentage of trips a driver has in each cluster, as indicated by the legend. Additionally, dashed lines in corresponding cluster colors mark the upper boundary of the average percentage of trips within each cluster.

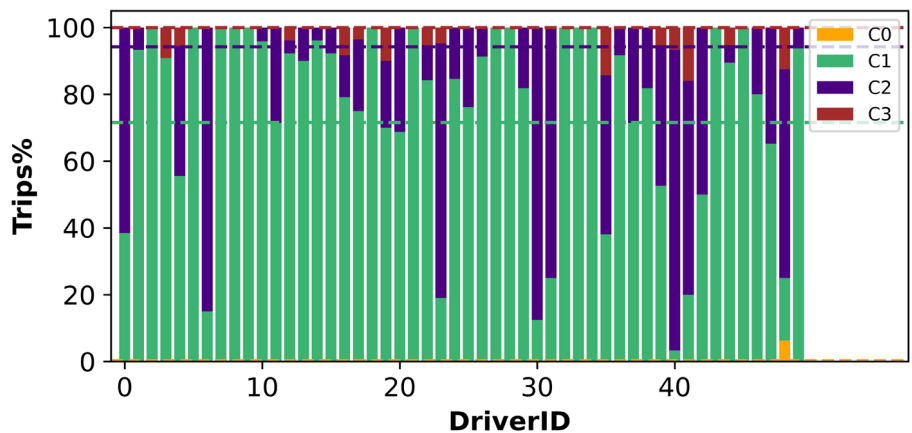

**Fig. 5.** Distribution of driver's fuel-efficiency performance in four clusters.

Fig. 5 indicates that the proportion of each driver's fuel-efficiency performance varies from the average trips' proportion across clusters. As Cluster 0 (the orange segment) contains insufficient samples for robust analysis, Clusters 1 to 3 (the green, purple, and brown segments respectively) can provide a detailed look into the drivers' fuel efficiency distributions. Significant differences between the segments' length of each bar and the position of the dashed lines of each cluster suggest notable differences in fuel efficiency among drivers.

The analysis highlights that individual driving behaviors significantly impact fuel efficiency, leading to considerable variances from the expected average distributions. In particular, the dominant segment proportion of Cluster 1 or 2 in individual drivers' bars underlines the influence of individual driving practices on fuel efficiency. In most individual trips, one cluster's trips comprise nearly 100%, while the remaining three clusters are almost 0%.

**Routes.** This section analyzes the fuel-efficiency performance of 44 routes in the recorded trips, based on the 'RouteID' feature from our dataset. The analysis categorizes the trips of different routes into four clusters in Section 4.1 to observe patterns in fuel efficiency. The trips of each route are assessed to determine their proportion across the four clusters. These percentages are then compared to the average proportion of all trips, to identify characteristics of routes in terms of fuel efficiency.

Fig. 6 demonstrates a bar chart illustrating trip proportions per route. Similar to the notations in Fig. 5, the x-axis lists the Routes' IDs, representing individual routes, and the y-axis shows the percentage of each route's executed trips in each cluster relative to their total trips. 'C0' – 'C3' in the legend denote 'Cluster 0' - 'Cluster 3' in Fig. 4 respectively. Each bar is segmented into four parts, reflecting the percentage of trips executed in each cluster, as indicated by the legend. Additionally, dashed lines in corresponding cluster colors indicate the upper boundary of the average percentage of trips within each cluster.

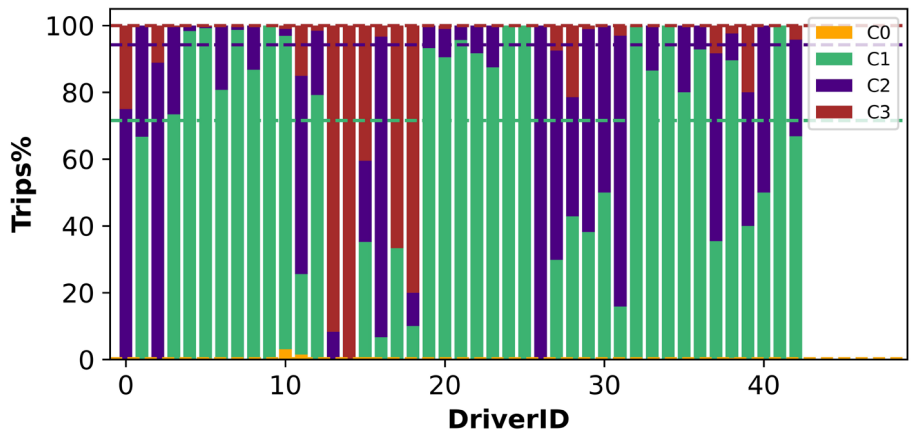

**Fig. 6.** Distribution of the fuel-efficiency performance in different routes in four clusters.

Fig. 6 indicates that the proportion of the fuel-efficiency performance in each route varies from the average trips' proportion across clusters. Similar to Fig. 5, Clusters 1 to 3 (the green, purple, and brown segments respectively) can offer a detailed examination of the routes' fuel efficiency distributions. The substantial differences between the lengths of the segments in each bar and the positions of the dashed lines for each cluster indicate notable variations in fuel efficiency across different routes.

The analysis highlights that individual route characteristics significantly impact fuel efficiency, leading to considerable variances from the expected average distributions. In particular, the dominant segment proportion of Cluster 1 or 2 in individual route bars underlines the influence of individual route characteristics on fuel efficiency. In most individual trips, one cluster's trips comprise nearly 100%, while the remaining three clusters are almost 0%.

## 5    Discussion and Future Research

**Clustering Performance.** Fig. 7 demonstrates the distribution of each cluster along with their outliers depicted through boxplots. The black hollow circles denote the outliers.

As seen in Fig. 7, samples in Cluster 1 and Cluster 2 are concentrated in a specific range without any outliers. However, Cluster 0 contains two outliers and Cluster 3 exhibits several outliers. This observation indicates that extreme fuel consumption is atypical in bus operations. Therefore, further investigation to understand their influencing factors should be conducted independently of the normal trips (Cluster 1&2). Furthermore, the recommendation of avoiding extremely low fuel efficiency could be explored as another topic for future research.

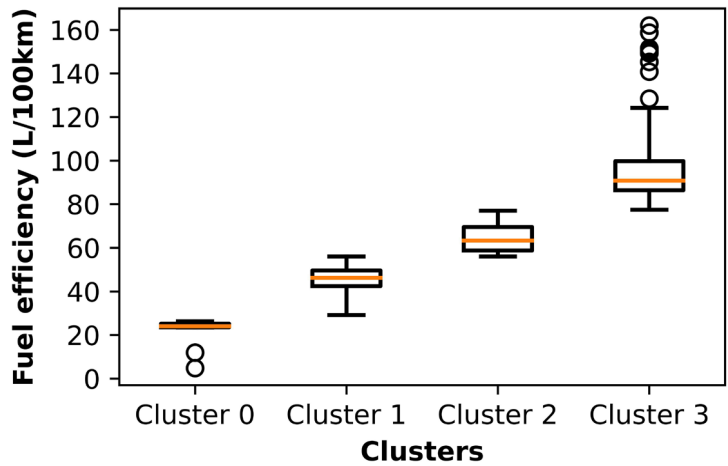

**Fig. 7.** Boxplots of each cluster.

**Fuel-efficiency Influencing Factors.** By examining the discrepancies in Fig. 5 and Fig. 6, it can be inferred that the fuel efficiency of distinct drivers and routes varies significantly. This variance underscores the substantial impact of individual driving habits and route characteristics. Consequently, personalized driver training and interventions are crucial to enhancing fuel efficiency across various routes. To provide more detailed recommendations on fuel efficiency enhancement, a thorough analysis of the relationship among features regarding driving behavior, route conditions, and fuel efficiency is necessary. For instance, trips characterized by low or extremely low fuel efficiency might benefit from modifications in driving practices or route adjustments. This analysis should involve correlation analysis and causal inference to elucidate the influencing factors of fuel efficiency.

## 6    Conclusion

To investigate the bus fuel consumption in public transportation systems, this paper employs the GMM clustering algorithm to categorize the fuel efficiency data and an integration method for optimal cluster selection. An integration method combining the Silhouette index, Calinski-Harabasz index, and Davies-Bouldin index is developed to select the optimal cluster numbers. Through the analysis of a dataset comprising 4006 bus trips in North Jutland, Denmark, four fuel efficiency categories are identified, including extreme fuel efficiency, normal fuel efficiency, low fuel efficiency, and extremely low fuel efficiency. Moreover, the significant impact of driving behaviors and route conditions on fuel efficiency is highlighted by the analysis of bar charts.

These findings emphasize the importance of targeted interventions to improve fuel efficiency in public transportation, thereby advancing sustainability goals and reducing environmental impact. Future studies should explore influential factors on fuel efficiency in each group, including correlation analysis and causal inference methodologies. Such comprehensive analyses could provide deeper insights into the factors affecting fuel consumption and support the development of more effective policies and practices in public transportation.

**Acknowledgments.** This work was funded by the Energy Technology Development and Demonstration Programme (EUDP) in Denmark under the project EUDP 2021-II Driver Coach [case no. 64021-2034].

**Disclosure of Interests.** The authors have no competing interests to declare that are relevant to the content of this article.